# Evolution of Convolutional Neural Network (CNN): Compute vs Memory bandwidth for Edge AI

*Dwith Chenna, Senior Embedded DSP Engineer, Computer Vision*
*MagicLeap Inc.* dchenna@magicleap.com

**Abstract—**Convolutional Neural Networks (CNNs) have greatly influenced the field of Embedded Vision and Edge Artificial Intelligence (AI), enabling powerful machine learning capabilities on resource-constrained devices. This article explores the relationship between CNN compute requirements and memory bandwidth in the context of Edge AI. We delve into the historical progression of CNN architectures, from the early pioneering models to the current state-of-the-art designs, highlighting the advancements in compute-intensive operations. We examine the impact of increasing model complexity on both computational requirements and memory access patterns. The paper presents a comparison analysis of the evolving trade-off between compute demands and memory bandwidth requirements in CNNs. This analysis provides insights into designing efficient architectures and potential hardware accelerators in enhancing CNN performance on edge devices.

## INTRODUCTION

Compute requirements of Artificial Intelligence (AI) models in computer vision (CV) has been increasing rapidly every year. This leads to hardware accelerators focused on computation, usually at the expense of removing other parts such as memory hierarchy. Many computes intense CNN applications have memory bandwidth communication as the bottleneck, in such accelerators. The DRAM memory accelerator scaling has been modest at 2-3x every 2 years, compared to the compute capabilities. The memory requirements for inference are usually much larger than the number of parameters. This is mainly due to the intermediate activation that requires 3-4x more memory, to move the data from/to on-chip local memory. These data transfer limitations can be between on-chip memory and DRAM memory or across different processors, where the bandwidth has been lagging significantly compared to the compute capabilities. As shown in Fig. 1. highlights the increase in compute capacity is 1500x when compared to the DRAM/interconnect bandwidth over the decade. These challenges are commonly referred to as the "memory wall" problem [8], which address both memory capacity and bandwidth.

Fundamental challenges of increasing DRAM/interconnect bandwidth [6], are difficult to overcome. This is intended to only increase the gap between compute and bandwidth capability, making it more challenging to deploy SOTA CNN models at the edge. Addressing this issue needs to rethink the
design CNN models, instead of simple scaling schemes based only on FLOPs or compute. The developments in hardware accelerators have been mainly focused on peak compute with limitations of the memory-bound bottleneck. This led to many CNN models that are bandwidth bound, resulting in inefficient utilization of these accelerators.

Model optimization techniques like pruning or quantization enable compressing these models for inference. Pruning is removal of redundant parameters in the model with minimal impact on accuracy. It is possible to prune up to 30% of the model with structured sparsity and 80% through unstructured sparsity with minimal impact of accuracy [15]. This is heavily dependent on the model architecture and any higher pruning results in significant accuracy degradation. Alternatively, quantization approaches have shown successful results in model compression. Quantization refers to the process of reducing the precision of the CNN models to FP16/INT8 or even low bit precision, which leads to significant reduction in memory footprint, bandwidth and latency [16]. INT8 quantization have shown successful results and are adopted by many popular open-source frameworks like tflite/pytorch. Due to the fundamental challenges in increasing compute and memory bandwidth, it is imperative that we rethink the architecture design and deployment of CNN models to handle these limitations. It is possible to sacrifice compute for better bandwidth performance resulting in models that led themselves to efficient deployment on the edge.

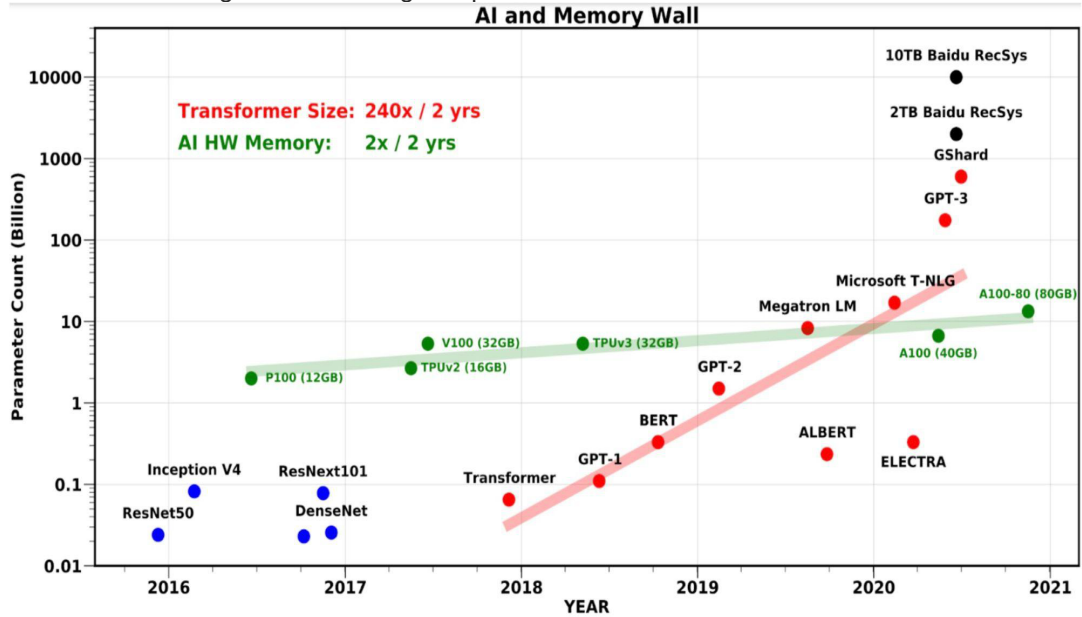

Fig. 1.  Scaling of computer and memory bandwidth for AI models [8]

In this paper, we will discuss the evolution of CNN architecture in Section 2.1, highlighting its implications on the compute and memory bandwidth. Section 2.2 and 2.3 we will discuss the underlying assumption for compute and memory bandwidth estimation. Section 3 will discuss the implementation of ONNX tools and analysis of results on a variety of CNN models.

## CONVOLUTIONAL NEURAL NETWORKS

Convolutional Neural Networks (CNN) have shown state of the art performance on several computer and Image processing challenges and datasets [1]. This led to CNNs being widely used in applications like Image classification, super resolution, segmentation and object detection. CNN has the ability to extract features at multiple stages, which allows them to learn data patterns especially for computer vision/image processing applications. Many experiments on using different activations, loss functions, operations, parameter optimization have been explored, which lead to increased representation capabilities through architecture innovations. CNN and their application in computer vision/image processing applications allows ideas of spatial, channel, depth of architecture, and multipath processing of information. Multilayer structure of CNN gives it ability to extract low and high-level features. These features can be transferred to generic tasks through Transfer Learning (TL) [1]. Many innovations in CNNs come from different aspects of network parameters, processing units, connectivity layers and optimization strategies. Recent developments in hardware technologies also allowed developing and deploying these models, making them ubiquitous in various applications running in cloud to edge applications.

In this section we focus on evolution of CNN architectures with emphasis on compute and memory bandwidth, allowing a much better understanding on the network architecture impact on deployment of the edge.

*AlexNet*
AlexNet, proposed by Krizhevsky et al., marked a significant advancement in Convolutional Neural Network (CNN) architectures for image classification and recognition tasks. It surpassed traditional methods by employing a deeper model and strategic parameter optimization [7]. To address hardware limitations, the model was trained on dual NVIDIA GTX 580 GPUs, allowing it to extend to 8 layers. This expansion enhanced its capacity to learn from diverse image categories and improved its adaptability to varying resolutions. Yet, deeper architectures brought the challenge of overfitting. With 60 million parameters, overfitting was mitigated through measures like dropout layers and data augmentation. The model's prowess was evident when it clinched the 2012 ImageNet competition, outperforming its closest competitor by a significant 11% in error reduction.

*VGG*
The success of CNNs in image recognition spurred architectural advancements. Simonyan et al.'s creation, VGG-16, employed simple yet effective principles [9] to enable deeper CNN models. This marked a significant leap in deep learning and computer vision, introducing the era of very deep CNNs. VGG-16 innovated by replacing larger filters with a compact 3x3 variant, showcasing that stacking smaller filters could replicate larger ones. This approach lowered computational complexity, reducing parameters and model size. The model's simplicity, depth, and uniform structure made it a frontrunner, securing 2nd place in the 2014 ILSVRC competition. However, its 138 million parameters posed computational challenges, limiting its deployment on edge devices.

*Inception*
Inception-V1, also known as GoogleNet, emerged as the victor of the 2014 ILSVRC competition with a focus on high accuracy and reduced computational complexity [10]. Named after its Inception blocks, the model employed multi-scale convolution transforms via split, transform, and merge techniques. These blocks encompassed filters of varying sizes (1x1, 3x3, and 5x4) to capture diverse spatial information, effectively addressing image category diversity at different resolutions. Computational efficiency was maintained through 1x1 convolution bottleneck layers preceding larger kernels, and a global average pool replaced the fully connected layer to reduce connection density. By applying such adjustments, parameters were pruned from 138 million to 4 million. This inception concept evolved in subsequent versions like Inception-V2, V3, V4, and Inception-ResNet, which introduced asymmetric filters to further streamline compute complexity.

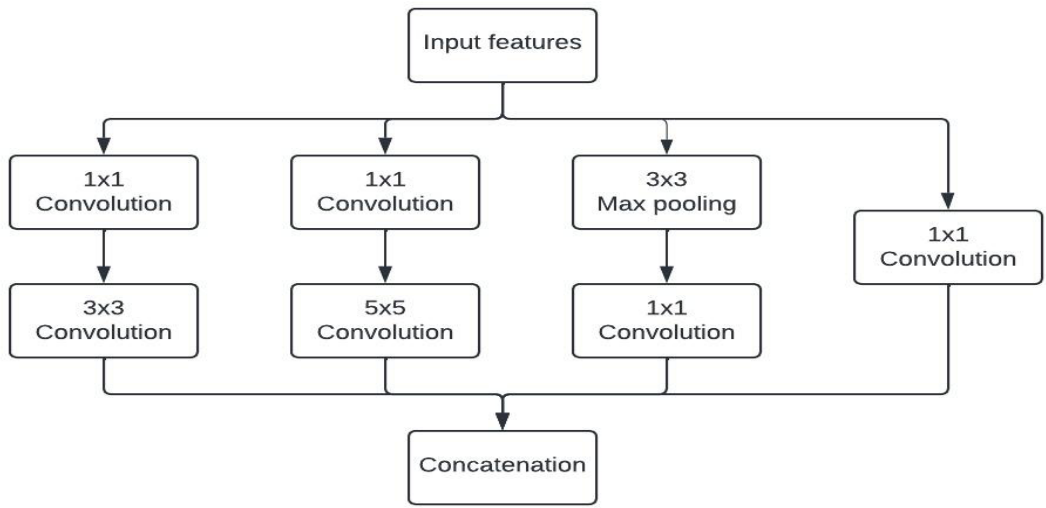

Fig. 2. Inception block showing the split compute and merge.

### ResNet

ResNet, devised by He et al. [5], brought a revolutionary shift in CNN architecture by introducing residual learning and efficient techniques for deep network training. It tackled the vanishing gradient issue, enabling even deeper CNN models. ResNet's innovation allowed a 152-layer deep CNN that triumphed in the 2015 ILSVRC competition. Compared to AlexNet and VGG, ResNet's 20x and 8x depth respectively came with relatively lower computational complexity. Empirical findings favored ResNet models with 50/101/152 layers over shallower variants. It showcased remarkable accuracy enhancements for complex visual tasks like image recognition and localization on the COCO dataset. ResNeXt [17] emerged as an improvement, treating it as an ensemble of smaller networks. Utilizing diverse convolutions (1x1, 3x3, 5x5) appended with 1x1 bottleneck convolution blocks, ResNeXt explored various topologies across different paths.

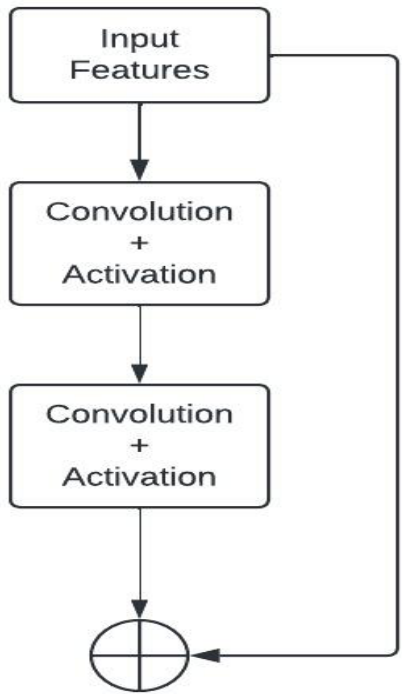

Fig. 3. Residual block structure used in ResNet

### DenseNet

DenseNet capitalizes on the insight that close connections between input and output lead to deeper, accurate, and efficient CNN models [18]. By linking each layer to the following layer in a feed-forward manner, DenseNet incorporates feature maps from all preceding layers through concatenation. This strategy enables the model to discern added and preserved information within the network. With a reasonable parameter count, the model utilizes a relatively smaller number of feature maps. Key attributes include significant parameter reduction, promotion of feature reuse, and handling of vanishing gradients. Impressively, DenseNet achieves performance on par with ResNet while demanding fewer parameters and computational resources.

### MobileNet

MobileNets [19] leverage depthwise convolutions to drastically decrease model size and computational demands. Engineered for resource-limited mobile devices with latency-sensitive tasks, the architecture allows flexibility through width and resolution multipliers. This empowers users to balance computation and latency as needed. The depthwise convolution based MobileNet reduces parameters from 29.3M to 4M and FLOPs by nearly 10x. These models, exemplified by MobileNet, achieve leaner and deeper structures, fostering representation diversity and computational efficiency. This design aligns well with cache-based systems and suits bandwidth-constrained edge AI applications due to their compact yet comprehensive nature.

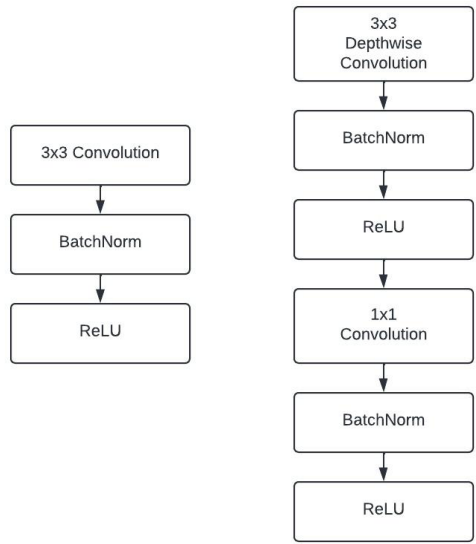

Fig. 4. Convolution layer with batchnorm and activation, depth wise convolution with 1x1 convolution layers

*SqueezeNet*

SqueezeNet stands out with its small footprint, boasting 50x fewer parameters yet achieving AlexNet-level accuracy on ImageNet [20]. Designed to enhance training communication and operate on resource-constrained hardware, it optimizes memory utilization. SqueezeNet's compactness arises from substituting 3x3 filters with 1x1 filters, resulting in a 9x parameter reduction. By employing squeeze layers, channel count reduction in 3x3 convolutions is utilized to maintain efficiency. Late stage downsampling, known as delayed downsampling [14], ensures higher accuracy is maintained.

*EfficientNet*

EfficientNet, developed through AutoML NAS [21], strategically optimizes accuracy and compute efficiency. Employing mobile inverted residual bottleneck convolutions (MBConv) akin to MobileNet, it leverages compound scaling [22] for diverse networks across computation budgets and model sizes. EfficientNet excels in accuracy and efficiency over existing CNNs, substantially reducing model size and FLOPs. For instance, EfficientNet-B0 surpasses ResNet-50's performance with a 5x parameter reduction and 10x FLOP reduction. These models outperform counterparts like ResNet, DenseNet, and Inception while employing significantly fewer parameters.

Table 1 summarizes the accuracy of these models on the ImageNet classification accuracy task. The overview of ImageNet accuracy across various network architectures offers insightful observations. Pioneers like AlexNet achieved commendable accuracy (57.2% top-1, 80.2% top-5), setting the foundation. VGG16 and VGG19 maintained steady performance at 71.5% top-1 accuracy. Inception models, like Inception V2 and GoogleNet, harnessed innovation for strong accuracy (up to 73.9%). The ResNet series pushed boundaries with deep structures, reaching a remarkable 78.3% top-1 accuracy. DenseNet's feature reuse led to 74.9% accuracy. Models like SqueezeNet (57.5%) optimized compactness, while MobileNet V2 (71.8%) and ShuffleNet V2 (67.9%) catered to mobile devices. EfficientNet achieved efficiency and accuracy (77.1%). Balancing accuracy, size, and computation remains crucial for diverse applications and hardware constraints.

| Model | Model Size | Top 1% | Top 5% |
| --- | --- | --- | --- |
| AlexNet | 238 MB | 54.80 | 78.23 |
| VGG16 | 527.8 MB | 72.62 | 91.14 |
| VGG19 | 508.5 MB | 73.72 | 91.58 |
| Inception V1 | 27 MB | 67.23 | 89.6 |
| Inception V2 | 44 MB | 73.9 | 91.8 |
| GoogleNet | 27 MB | 67.78 | 88.34 |
| ResNet18 | 44.7 MB | 69.93 | 89.29 |
| ResNet34 | 83.3 MB | 73.73 | 91.40 |
| ResNet50 | 97.8 MB | 74.93 | 92.38 |
| ResNet101 | 170.6 MB | 76.48 | 93.20 |
| ResNet152 | 230.6 MB | 77.11 | 93.61 |
| DenseNet | 32 MB | 60.96 | 92.2 |
| SqueezeNet | 5 MB | 56.85 | 79.87 |
| MobileNet V2 | 13.3 MB | 69.48 | 89.26 |
| ShuffleNet V2 | 8.79MB | 66.35 | 86.57 |
| EfficientNet | 51.9 MB | 80.4 | 93.6 |

Table 1. Summary of Image classification accuracy results for different mode

## COMPUTE

As CNN architectures become deeper and more complex, it's essential to estimate their computational requirements accurately. Two commonly used metrics for estimating computational capacity are Floating Point Operations per Second (FLOPs) and Multiply-Accumulate (MAC) operations. FLOPs provide an estimate of the total number of floating-point operations (additions and multiplications) required to perform the forward pass of a neural network. For CNNs, this includes operations like convolutions, pooling, element wise addition/multiplication and fully connected layers. MAC operations are a fundamental building block of many mathematical computations, especially in matrix multiplications. In the context of CNNs, MAC operations occur during convolutional and fully connected layers. For operations like convolution and fully connected layers, we can estimate FLOPs and MAC through

$FLOP \simeq 2 * MAC$

In this section we estimate the compute complexity of the CNN through FLOPs, which can better estimate operations like elementwise add/multiply, average pooling which are found more frequently in many of these modern architectures. Estimation of FLOPs for different architectures is dependent on the different operations used in the network architecture. We will discuss widely used operations and how to estimate the FLOPs for these operations. Due to the simplified structure and scaling of the deep CNNs model we tend to see many repetitive operations across different network architecture. This makes it easy to estimate the compute complexity by analyzing different operations used to form the layer of the network. This section section we will review five major types of operations and estimate FLOPs for these operations.

### Convolution

Convolutions is the fundamental building block for CNN models. It leverages the spatial sparsity to reduce the number of operations and still be effective in image processing/computer vision applications. It usually consists of multi-dimensional input and weight that generate the output feature maps as shown in Fig. 5.

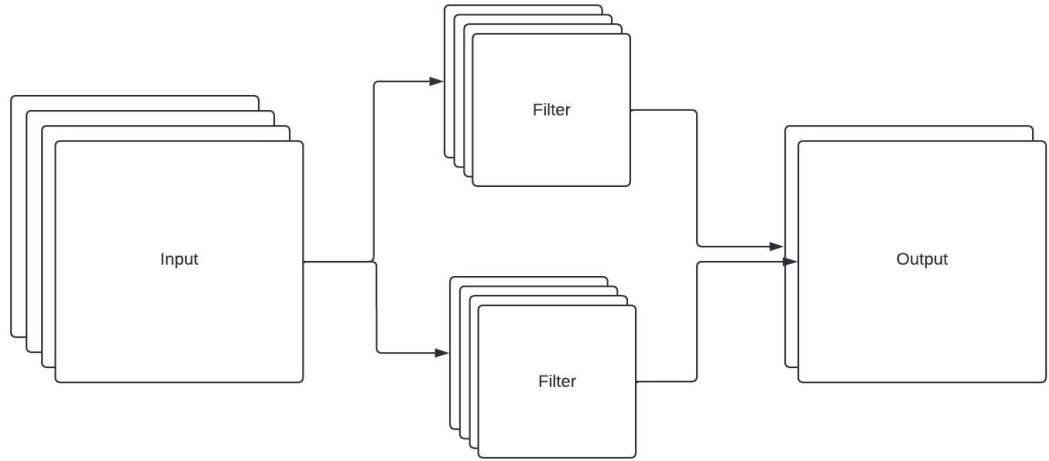

Fig. 5. Convolution operation with multiple filters

The convolutions operations can be represented through the equation (1), we can estimate the FLOPs based on the input, weight and output dimensions.

$$y_{out}^{i,k} = \sum_{i,j=1}^{M} x_{in}^{i,j} w^{i,j} + b^k \qquad (1)$$

flops = C_in * H_out * W_out * C_out * H_k * W_k / (stride * stride * group)   (2)

Where C_in is input channel dimension, H/W/C_out are output dimensions, H/W_k kernel dimensions, stride and group.

### Fully Connected (FC)

Fully connected layers are usually seen at the head of the network. Due to their extensive connectivity they tend to have orders of magnitude compute that increase with the fixes. In most cases the feature map sizes are reduced to avoid such bottlenecks in the network.

flops = N_in * H_w * W_w   (3)

### Average Pool

Spatial dimensions of the feature maps are reduced as we go deeper into the network, allowing the network to capture high spatial features. This is achieved by using pooling operations that reduce the spatial dimension with fixed kernel size and stride. Average or max pooling operations are widely used in CNNs. In case of max pooling operations, the computation is negligible as it picks the max from the kernel dimensions. In this section we will look at estimating compute for average pooling operation.

flops = C_in * H_out * W_out * H_k * W_k / stride   (4)

Where C_in is input channel dimension, H/W_out are output feature map dimensions, H/W_k are the pooling kernel dimensions and stride.

### Elementwise Add/Mul

Elementwise Add/Multiply operations are prevalent in many modern network architectures after ResNet architecture showed promising results to vanishing gradient problem. This enabled deeper networks with larger representation capacity. The compute estimate for these operations is through eq (5), where C_in are input feature dimensions, H/W_in are input spatial dimensions.

flops = C_in * H_in * W_in   (5)

## MEMORY BANDWIDTH

Memory bandwidth is a critical consideration when deploying Convolutional Neural Network (CNN) models for inference on hardware platforms. CNNs involve intensive data movement between memory and processing units, making memory bandwidth a potential bottleneck for performance. Estimating memory bandwidth requirements helps in optimizing model deployment and selecting appropriate hardware. Memory bandwidth refers to the rate at which data can be read from or written to memory. In the context of CNN inference, memory bandwidth is crucial because it involves data movement that needs to be moved between memory and processing unit. Two main components of data movement include i) Weight Loading, CNN layers utilize learned weights that are stored in memory. Fetching these weights efficiently is vital for smooth inference ii) Feature Maps, Intermediate feature maps generated during computation are stored in memory for subsequent layers' use. Efficient access to these maps impacts inference speed.

Edge AI devices usually remove memory hierarchy in favor of larger compute, small silicon area and power. These systems have dedicated I/O control blocks that allow instruction/data to move independently. Independent instruction/data buses allow users to hide the memory and compute latency through a double buffer, which increases the overall throughput. These architectures are memory and power efficient making them a popular choice for resource constrained hardware with strict latency and power budget.

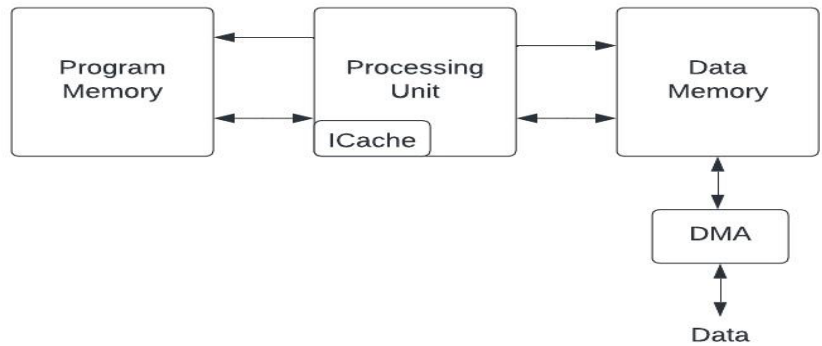

Fig. 6. Super Harvard Architecture improvement include instruction cache and dedicated DMA controller.

In order to estimate bandwidth, we need to move the intermediate activation to and from the local memory to system memory. This puts extra burden on the bandwidth requirement, when estimating bandwidth, we need to consider the data from system memory to local memory and back, resulting in twice the memory movement. These estimates are made per model inference, by calculating the intermediate activation and weights that need to be moved in and out of local and system memory.

## IMPLEMENTATION

In order to estimate the computer and memory bandwidth requirements, based on discussion we use ONNX models. Open Neural Network Exchange (ONNX) [14], is an open-source machine independent format, widely used for exchanging neural network models. We infer the sizes of intermediate results for the ONNX model in order to estimate the bandwidth requirement. For bandwidth estimation we assume model optimization techniques batchnorm folding and layer fusion, which fuses batch normalization and activation functions into the previous convolution operation.

## RESULTS

In this section, we will review the results of compute and memory estimation. These estimates are based on a group of CNN models that represent the evolution of CNN architectures throughout the past few years. For this analysis we used AlexNet[7], VGG-16[9], InceptionV1/V2[10], ResNet18-152[5],

DenseNet[18], MobileNet[19], SqueezeNet[20] and EfficientNet[22], as discussed in Section 2. Fig. 7. shows network vs compute (FLOPs), the FLOPs are in log scale to accommodate a variety of networks. All these ONNX models [] use standard input sizes of 224x224x3, which enables us to do a fair comparison across different architectures.
`

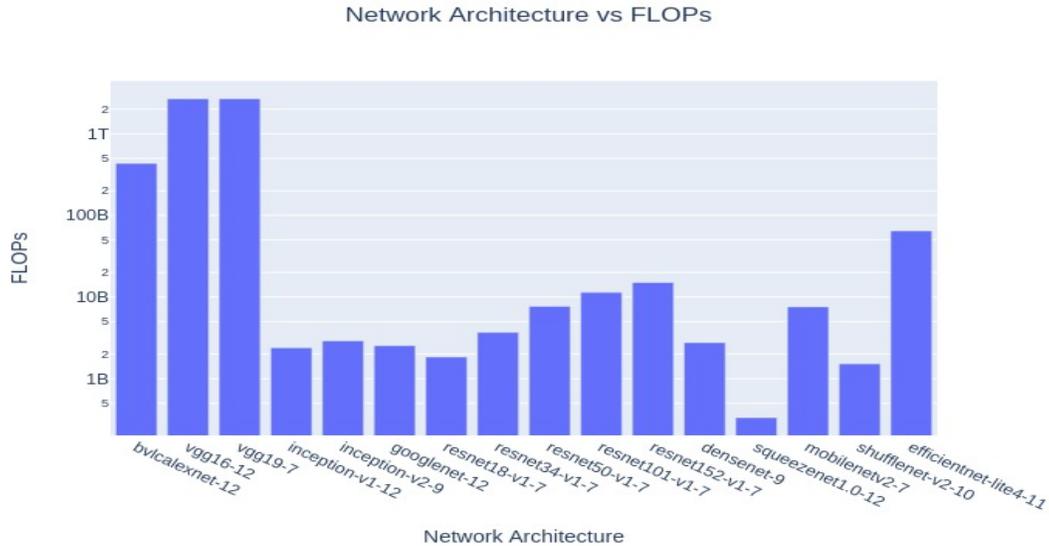

Fig. 7. Network architecture vs Compute capacity (FLOPs)

We clearly see a trend towards small FLOPs, as the network architecture matures with different architecture optimization techniques. However, bandwidth doesn't show any consistent trend, it is from the fact that many of these architecture designs do not consider the bandwidth constraint of the edge AI applications.

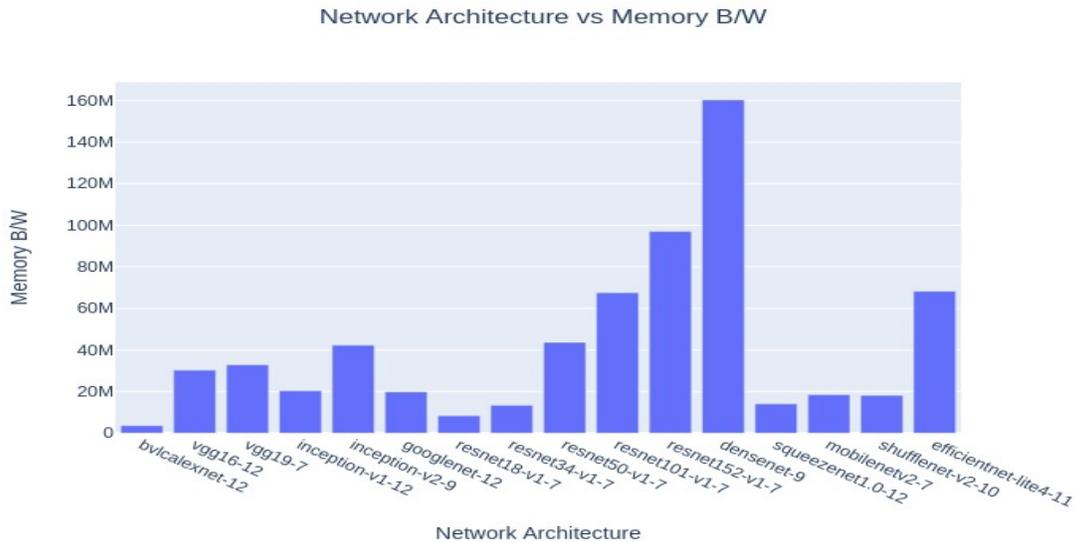

Fig. 8. Network architecture vs Memory Bandwidth (/Inference)

A comparison of the compute vs memory bandwidth (Fig. 9), shows that mobilenet networks designed for mobile application tend to be leaner and deeper, requiring higher bandwidth, might not be optimal architecture for the edge AI application. SqueezeNet shows superior performance in terms of both compute and memory bandwidth tradeoff, as it was designed for smaller size and minimal memory bandwidth.

## SUMMARY

This paper provides a comprehensive analysis of the evolution of CNN compute requirements and memory bandwidth in Edge AI applications. The rapid evolution of CNNs models, using FLOPs/parameters as estimates of performance results in suboptimal models. Especially hardware bandwidth constraints act as bottlenecks to model performance. It is essential to estimate both compute and bandwidth to design architecture that provides optimal performance for Edge AI application, with strict constraints on latency, bandwidth and power. In this paper, we offer valuable insights to researchers, practitioners, and hardware designers, facilitating a deeper understanding of the trade-offs involved in optimizing CNNs for edge deployment.

## CONFLICT OF INTEREST

The author of the manuscript hereby declares that we have no conflicts of interest to disclose related to the research work presented in this manuscript. I confirm that no financial, personal, or professional relationships or affiliations exist that could be perceived as a conflict of interest with regards to the research presented in the manuscript. I further confirm that no funding sources or sponsors played a role in the design, execution, analysis, or interpretation of the study or in the preparation of this manuscript.

## DATA AVAILABILITY

The datasets generated during and/or analyzed during the current study are available in the github repository: https://github.com/onnx/models/tree/main/vision/classification.
The tools used for the analysis will be made available in github repository:
https://github.com/cyndwith/Evolution-of-Convolutional-Neural-Network-CNN-Compute-vs-Memory-bandwidth-for-Edge-AI

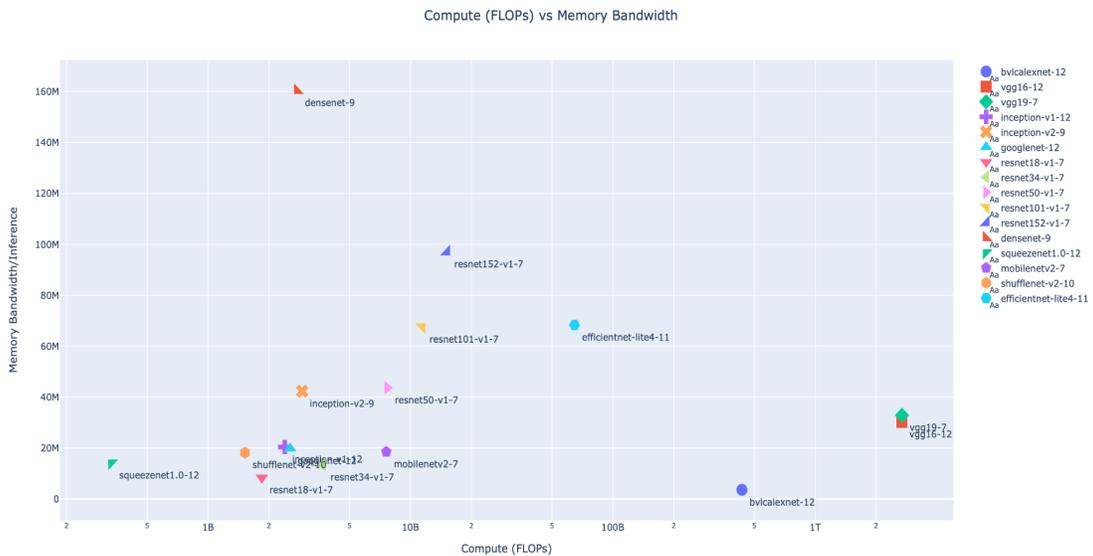

Fig. 9. Computer (FLOPs) vs Memory Bandwidth (/Inference)